# An Online Learning-based Framework for Tracking

**Kamalika Chaudhuri**
Computer Science and Engineering
University of California, San Diego
La Jolla, CA 92093

**Yoav Freund**
Computer Science and Engineering
University of California, San Diego
La Jolla, CA 92093

**Daniel Hsu**
Computer Science and Engineering
University of California, San Diego
La Jolla, CA 92093

## Abstract

We study the tracking problem, namely, estimating the hidden state of an object over time, from unreliable and noisy measurements. The standard framework for the tracking problem is the generative framework, which is the basis of solutions such as the Bayesian algorithm and its approximation, the particle filters. However, these solutions can be very sensitive to model mismatches. In this paper, motivated by online learning, we introduce a new framework for tracking. We provide an efficient tracking algorithm for this framework. We provide experimental results comparing our algorithm to the Bayesian algorithm on simulated data. Our experiments show that when there are slight model mismatches, our algorithm outperforms the Bayesian algorithm.

## 1 INTRODUCTION

We study the tracking problem, which has numerous applications in AI, control and finance. In tracking, we are given noisy measurements over time, and the problem is to estimate the hidden state of an object. The challenge is to do this reliably, by combining measurements from multiple time steps and prior knowledge about the state dynamics, and the goal of tracking is to produce estimates that are as close to the true states as possible.

The most popular solutions to the tracking problem are the Kalman filter (Kalman, 1960), the particle filter (Doucet et al., 2001), and their numerous extensions and variations (*e.g.* (Isard & Blake, 1998; van der Merwe et al., 2000)), which are based on a generative framework for the tracking problem. Suppose we want to track the state $x_t$ of an object at time $t$, given only measurement vectors $M(\cdot, t')$ for times $t' \leq t$. In the generative approach, we think of the state $X(t)$ and measurements $M(\cdot, t)$ as random variables. We represent our knowledge regarding the dynamics of the states using the transition process $\Pr(X(t)|X(t-1))$ and our knowledge regarding the (noisy) relationship between the states and the observations by the measurement process $\Pr(M(\cdot, t)|X(t))$. Then, given only the observations, the goal of tracking is to estimate the hidden state sequence $(x_1, x_2, \ldots)$. This is done by calculating the likelihood of each state sequence and then using as the estimate either the sequence with the highest posterior probability (maximum a posteriori) or the expected value of the state with respect to the posterior distribution (the Bayesian algorithm). In practice, one uses particle filters, which are an approximation to the Bayesian algorithm.

The problem with the generative framework is that in practice, it is very difficult to precisely determine the distributions of the measurements. The Bayesian algorithm can be sensitive to model mismatches, so using a model which is different from that generating the measurements can lead to a large divergence between the estimated states and the true states.

To address this, we introduce an online learning-based framework for tracking. In our framework, we are given a set of state sequences or paths in the state space; but instead of assuming that the observations are generated by a measurement model from a path in this set, we think of each path as a mechanism for explaining the observations. We emphasize that this is done regardless of how the observations are generated. Suppose a path $(x_1, x_2, \ldots)$ is proposed as an explanation of the observations $(M(\cdot, 1), M(\cdot, 2), \ldots)$. We measure the quality of this path using a predefined *loss function*, which depends only on the measurements (and not on the hidden true state). The tracking algorithm selects its own path by taking a weighted average of the best paths according the past observations. The theoretical guarantee we provide is that the loss of the path selected by the algorithm in this online way by the tracking algorithm is close to that of the path with the minimum loss; here, the loss is measured according to the loss function supplied to the algorithm. Such guarantees are analogous to competitive analysis used in online learning (Cesa-Bianchi & Lugosi, 2006; Freund & Schapire, 1997; Littlestone & Warmuth, 1994), and it is important to note that such guarantees hold uniformly for *any* sequence of observations, regardless of any probabilistic assumptions.

Our next contribution is to provide an online learning-based algorithm for tracking in this framework. Our algorithm is based on NormalHedge (Chaudhuri et al., 2009), which is a general online learning algorithm. NormalHedge can be instantiated with *any loss function*. When supplied with a bounded loss function, it is guaranteed to produce a path with loss close to that of the path with the minimum loss, from a set of candidate paths. As it is computationally inefficient to directly apply NormalHedge to tracking, we derive a Sequential Monte Carlo approximation to NormalHedge, and we show that this approximation is efficient.

To demonstrate the robustness of our tracking algorithm, we perform simulations on a simple one-dimensional tracking problem. We evaluate tracking performance by measuring the average distance between the states estimated by the algorithms, and the true hidden states. We instantiate our algorithm with a simple clipped loss function. Our simulations show that our algorithm consistently outperforms the Bayesian algorithm, under high measurement noise, and a wide range of levels of model mismatch.

We finally note that Bayesian algorithm can also be interpreted in an online learning-based framework. In particular, if the loss of a path is the negative log-likelihood (the log-loss) under some measurement model, then, the Bayesian algorithm can be shown to produce a path with log-loss close to that of the path with the minimum log-loss. One may be tempted to think that our tracking solution follows the same approach; however, the point of our paper is that one can use loss functions that are different from log-loss, and in particular, we show a scenario in which using other loss functions produces better tracking performance than the Bayesian algorithm (or its approximations).

The rest of the paper is organized as follows. In Section 2, we explain in detail our explanatory model for tracking. In Section 3, we present NormalHedge, on which our tracking algorithm is based. In Section 4, we provide our tracking algorithm. Section 5 presents the experimental comparison of our algorithm with the Bayesian algorithm on simulated data. Finally, we discuss related work in Section 6.

## 2 AN ONLINE-LEARNING FRAMEWORK FOR TRACKING

In this section, we describe in more detail the setup of the tracking problem, and our online learning-based framework for tracking. In tracking, at each time $t$, we are given as input, measurements (or observations) $M(\cdot, t)$, and the goal is to estimate the hidden state of an object using these measurements, and our knowledge about the state dynamics.

In our online learning framework for tracking, we are given a set $\mathcal{P}$ of paths (sequences) over the state space $\mathcal{X} \subset \mathbb{R}^n$. At each time $t$, we assign to each path in $\mathcal{P}$ a loss function $\ell$. The loss function has two parts: a dynamics loss $\ell_d$ and an observation loss $\ell_o$.

The dynamics loss $\ell_d$ captures our knowledge about the state dynamics. For simplicity, we use a dynamics loss $\ell_d$ that can be written as

$$\ell_d(\mathbf{p}) = \sum_t \ell_d(x_t, x_{t-1})$$

for a path $\mathbf{p} = (x_1, x_2, \ldots)$. In other words, the dynamics loss at time $t$ depends only on the states at time $t$ and $t-1$. A common way to express our knowledge about the dynamics is in terms of a dynamics function $F$, defined so that paths with $x_t \approx F(x_{t-1})$ will have small dynamics loss.

For example, consider an object moving with a constant velocity. Here, if the state $x_t = (p, v)$, where $p$ is the position and $v$ is the velocity, then we would be interested in paths in which $x_t \approx x_{t-1} + (v, 0)$. In these cases, the dynamics loss $\ell_d(x_t, x_{t-1})$ is typically a growing function of the distance from $x_t$ to $F(x_{t-1})$.

The second component of the loss function is an observation loss $\ell_o$. Given a path $\mathbf{p} = (x_1, x_2, \ldots)$, and measurements $\mathbf{M} = (M(\cdot, 1), M(\cdot, 2), \ldots)$, the observation loss function $\ell_o(\mathbf{p}, \mathbf{M})$ quantifies how well the path $\mathbf{p}$ explains the measurements. Again, for simplicity, we restrict ourselves to loss functions $\ell_o$ that can be written as:

$$\ell_o(\mathbf{p}, \mathbf{M}) = \sum_t \ell_o(x_t, M(\cdot, t)) \ .$$

In other words, the observation loss of a path at time $t$ depends only on its state at time $t$ and the measurements at time $t$. The total loss of a path $\mathbf{p}$ is the sum of its dynamics and observation losses. We note that the loss of a path depends only on that particular path and the measurements, and not on the true hidden state. As a result, the loss of a path can always be computed by an algorithm at any given time.

The algorithmic framework we consider in this model is analogous to, and motivated by the decision-theoretic framework for online learning (Freund &

Schapire, 1997; Cesa-Bianchi & Lugosi, 2006). At time $t$, our algorithm assigns a weight $w_{\mathbf{p}}^t$ to each path $\mathbf{p}$ in $\mathcal{P}$. The estimated state at time $t$ is the weighted mean of the states, where the weight of a state is the total weight of all paths in this state. The loss of the algorithm at time $t$ is the weighted loss of all paths in $\mathcal{P}$. The theoretical guarantee we look for is that the loss of the algorithm is close to the loss of the best path in $\mathcal{P}$ in hindsight (or, close to the loss of the top $\epsilon$-quantile path in $\mathcal{P}$ in hindsight). Thus, if $\mathcal{P}$ has a small fraction of paths with low loss, and if the loss functions successfully capture the tracking performance, then, the sequence of states estimated by the algorithm will have good tracking performance.

**Algorithm 1** NormalHedge

**initialize** $R_{i,0} = 0$, $w_{i,1} = 1/N$ $\forall i$
   **for** $t = 1, 2, \ldots$ **do**
      Each action $i$ incurs loss $\ell_{i,t}$.
      Learner incurs loss $\ell_{A,t} = \sum_{i=1}^N w_{i,t} \ell_{i,t}$.
      Update cumulative regrets: $R_{i,t} = R_{i,t-1} + (\ell_{A,t} - \ell_{i,t})$ $\forall i$.
      Find $c_t > 0$ satisfying
      $\frac{1}{N} \sum_{i=1}^N \exp\left(\frac{([R_{i,t}]_+)^2}{2c_t}\right) = e$.
      Update distribution:
      $w_{i,t+1} \propto \frac{[R_{i,t}]_+}{c_t} \exp\left(\frac{([R_{i,t}]_+)^2}{2c_t}\right)$ $\forall i$.
   **end for**

## 3 NORMALHEDGE

In this section, we describe the NormalHedge algorithm, which forms the basis of our tracking algorithm. To present NormalHedge in full generality, we first need to describe the decision-theoretic framework for online learning.

The problem of decision-theoretic online learning is as follows. At each round, a learner has access to a set of $N$ *actions*; for our purposes, an action is any method that provides a prediction in each round. The learner maintains a distribution $w_{i,t}$ over the actions at time $t$. At each time period $t$, each action $i$ suffers a loss $\ell_{i,t}$ which lies in a bounded range, and the loss of the learner is $\sum_i w_{i,t} \ell_{i,t}$. We note that this framework is very general – no assumption is made about the nature of the actions and the distribution of the losses. The goal of the learner is to maintain a distribution over the actions, such that its cumulative loss over time is low, compared to the cumulative loss of the action with the lowest cumulative loss. In some cases, particularly, when the number of actions is very large, we are interested in achieving a low cumulative loss compared to the top $\epsilon$-*quantile* of actions. Here, for any $\epsilon$, the top $\epsilon$-quantile of actions are the $\epsilon$ fraction of actions which have the lowest cumulative loss.

Starting with the seminal work of Littlestone and Warmuth (1994), the problem of decision-theoretic online learning has been well-studied in the literature (Freund & Schapire, 1997; Cesa-Bianchi et al., 1993; Cesa-Bianchi & Lugosi, 2006). The most common algorithm for this problem is Hedge or Exponential Weights (Freund & Schapire, 1997), which assigns to each action a weight exponentially small in its total loss. In this paper however, we consider a different algorithm NormalHedge for this problem (Chaudhuri et al., 2009), and it is this algorithm that forms the basis of our tracking algorithm. While the Bayesian averaging algorithm can be shown to be a variant of Hedge when the loss function is the log-loss, such is not the case for NormalHedge, which uses a very different weighting function.

In the NormalHedge algorithm, for each action $i$ and time $t$, we use $w_{i,t}$ to denote the *NormalHedge weight* assigned to action $i$ at time $t$. At any time $t$, we define the regret $R_{i,t}$ of our algorithm to an action $i$ as the difference between the cumulative loss of our algorithm and the cumulative loss of this action. Also, for any real number $x$, we use the notation $[x]_+$ to denote $\max(0, x)$. The NormalHedge algorithm is presented below. The performance guarantees for the NormalHedge algorithm can be stated as follows.

**Theorem 1** (Chaudhuri et al., 2009). *If NormalHedge has access to $N$ actions, then for all loss sequences, for all $t$, for all $0 < \epsilon \leq 1$, the regret of the algorithm to the top $\epsilon$-quantile of the actions is $O(\sqrt{t \cdot \ln(1/\epsilon)} + \ln^2 N)$.*

A significant advantage of using NormalHedge over other online learning algorithms is that it has no parameters to tune, yet achieves performance comparable to the best performance of previous online learning algorithms with optimally tuned parameters. For more discussion, see Section 3 of (Chaudhuri et al., 2009).

Another advantage is that the actions which have total loss greater than the total loss of the algorithm, are assigned zero weight. Since the algorithm performs almost as well as the best action, in a scenario where a few actions are significantly better than the rest, the algorithm will assign zero weight to most actions. In other words, the support of the NormalHedge weights may be a very small set, and this property makes it easier to design an approximation to it.

# 4 TRACKING USING NORMALHEDGE

To apply NormalHedge directly to tracking, we set each action to be a path in the state space, and the loss of each action at time $t$ to be the loss of the corresponding path at time $t$. To make NormalHedge more robust in a practical setting, we make a small change to the algorithm: instead of using cumulative loss, we use a discounted cumulative loss. For a discount parameter $0 < \alpha < 1$, the discounted cumulative loss of an action $i$ at time $T$ is $\sum_{t=1}^{T}(1-\alpha)^{T-t}\ell_{i,t}$. Such discounted losses have been used in reinforcement learning (Kaelbling et al., 1996) as well as online learning (Hazan & Seshadhri, 2009); intuitively, it makes the tracking algorithm more flexible, and allows it to more easily recover from past mistakes.

However, a direct application of NormalHedge is prohibitively expensive in terms of computation. If we consider paths over a discretization of the state space of cardinality $S$, then, at time $T$, there are $S^T$ actions. One can take advantage of the structure of the loss function to formulate the weight updates as a dynamic program; however, this is still expensive as each update takes $S^2$ time. Therefore, in the sequel, we show how to derive a Sequential Monte-Carlo based approximation to NormalHedge, and we use this approximation in our experiments.

The key observation behind our approximation is that the weights on actions generated by the NormalHedge algorithm induce a distribution over the states at each time $t$. We therefore use a random sample of states in each round to approximate this distribution. Thus, just as particle filters approximate the posterior density on the states induced by the Bayesian algorithm, our tracking algorithm approximates the distribution induced on the states by NormalHedge for tracking.

The main difference between NormalHedge and our approximation is that while NormalHedge always maintains the weights for all the actions, we delete an action from our action list when its weight falls to 0. We then replace this action by our re-sampling procedure, which chooses another action which is currently in a region of the state space where the actions have low losses. Thus, we do not spend resources maintaining and updating weights for actions which do not perform well. Another difference between NormalHedge and our tracking algorithm is that in our approximation, we do not explicitly impose a dynamics loss on the actions. Instead, we use a re-sampling procedure that only considers actions with low dynamics loss. This also avoids spending resources on actions which have high dynamics loss anyway. We note that because of these changes, our tracking algorithm does not have the worst-case theoretical guarantees in Theorem 1; however, we still expect it to have good performance when tracking a slowly moving object.

**Algorithm 2** Tracking algorithm
**input** $N$ (number of actions), $\alpha$ (discount factor), $\Sigma_*$ (re-sampling parameter), $F$ (dynamics function)
$\mathcal{A} := \{x_{1,1}, \ldots, x_{N,1}\}$ with $x_{i,1}$ randomly drawn from $\mathcal{X}$; $R_{i,0} := 0$; $w_{i,0} := 1/N$ $\forall i$
**for** $t = 1, 2, \ldots$ **do**
  Obtain losses $\ell_{i,t} = \ell_o(x_{i,t})$ for each action $i$ and update regrets:
  $R_{i,t} := (1-\alpha)R_{i,t-1} + (\ell_{A,t} - \ell_{i,t})$ where $\ell_{A,t} = \sum_{i=1}^N w_{i,t-1}\ell_{i,t}$.
  Delete poor actions: let $X = \{i : R_{i,t} \leq 0\}$, set $\mathcal{A} := \mathcal{A} \setminus X$.
  Re-sample actions:
  $\mathcal{A} := \mathcal{A} \cup \mathsf{Resample}(X, \Sigma_*, t)$.
  Compute weight of each action $i$: $w_{i,t} \propto \frac{[R_{i,t}]_+}{c} \exp\left(\frac{([R_{i,t}]_+)^2}{2c}\right)$
  where $c$ is the solution to the equation $\frac{1}{N}\sum_{i=1}^N \exp\left(\frac{([R_{i,t}]_+)^2}{2c}\right) = e$.
  Estimate: $x_{A,t} := \sum_{i=1}^N w_{i,t}x_{i,t}$.
  Update states: $x_{i,t+1} := F(x_{i,t})$ $\forall i$.
**end for**

Our tracking algorithm is specified in Algorithm 2. Each action $i$ in our algorithm is a path $(x_{i,1}, x_{i,2}, \ldots)$ in the state space $\mathcal{X} \subset \mathbb{R}^n$. However, we do not maintain this entire path explicitly for each action; rather, the state update step of the algorithm computes $x_{i,t+1}$ from $x_{i,t}$ using the dynamics function $F$, so we only need to maintain the current state of each action. Recall, applying the dynamics function $F$ should ensure that the path incurs no or little dynamics loss (see Section 2).

We start with a set of actions $\mathcal{A}$ initially positioned at states uniformly distributed over the $\mathcal{X}$, and a uniform weighting over these actions. In each round, like NormalHedge, each action incurs a loss determined by its current state, and the tracker incurs the expected loss determined by the current weighting over actions. Using these losses, we update the cumulative (discounted) regrets to each action. However, unlike NormalHedge, we then delete all actions with zero or negative regret, and replace them using a re-sampling procedure. This procedure replaces poorly performing actions with actions currently at high density regions of $\mathcal{X}$, thereby providing a better approximation to the NormalHedge weights.

The re-sampling procedure is explained in detail in Algorithm 3. The main idea is to sample from the regions of the state space with high weight. This is done

**Algorithm 3** Re-sampling algorithm

**input** $X$ (actions to be re-sampled), $\Sigma_*$ (re-sampling parameter), $t$ (current time)
   **for** $j \in X$ **do**
      Set $\bar{X} := \{i : R_{i,t} > 0\}$.
      If $\bar{X} = \emptyset$: set $\bar{p}_i = 1/N\ \forall i$. Else: set $\bar{p}_i \propto w_{i,t-1}$ $\forall i \in \bar{X}$ and $\bar{p}_i = 0\ \forall i \notin \bar{X}$.
      Draw $i \sim \text{Multinomial}(\bar{p}_1, \ldots, \bar{p}_N)$.
      Draw $x_{j,t} \sim \mathcal{N}(x_{i,t}, \Sigma_*)$, and set $R_{j,t} := (1 - \alpha) R_{i,t-1} + (\ell_{A,t} - \ell_o(x_{j,t}))$.
   **end for**

by sampling an action proportional to its weight in the previous round. We then choose a state randomly (roughly) from an ellipsoid $\{x : (x - x_t)^\top \Sigma_*^{-1} (x - x_t) \leq 1\}$ around the current state $x_t$ of the selected action; the new action inherits the history of the selected action, but has a current state which is different from (but close to) the selected action. This latter step makes the new state distribution smoother than the one in the previous round, which may be supported on just a few states if only a few actions have low losses. We note that $\Sigma_*$ can be set so that the re-sampling procedure only samples actions with low dynamics loss (and the state update step of the algorithm ensures that the remaining actions in the set $\mathcal{A}$ do not incur any dynamics loss); thus, our algorithm does not explicitly compute a dynamics loss for each action.

## 5 SIMULATIONS

For our simulations, we consider the task of tracking an object in a simple, one-dimensional state space. To evaluate our algorithm, we measure the distance between the estimated states, and the true states of the object.

Our experimental setup is inspired by the application of tracking faces in videos, using a standard face detector (Viola & Jones, 2001). In this case, the state is the location of a face, and each measurement corresponds to a score output by the face detector for a region in the current video frame. The goal is to predict the location of the face across several video frames, using these scores produced by the detector. The detector typically returns high scores for several regions around the true location of a face, but it may also erroneously produce high scores elsewhere. And though in some cases the detection score may have a probabilistic interpretation, it is often difficult to accurately characterize the distribution of the noise.

The precise setup of our simulations is as follows. The object to be tracked remains stationary or moves with velocity at most 1 in the interval $[-500, 500]$. At time

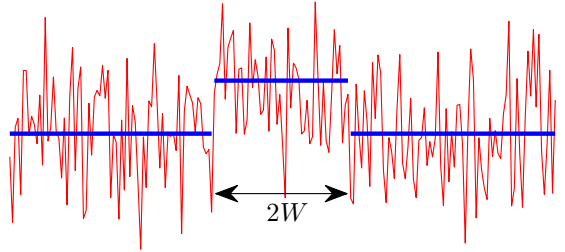

Blue: $H(x, z_t)$, Red: $M(x, t)$.

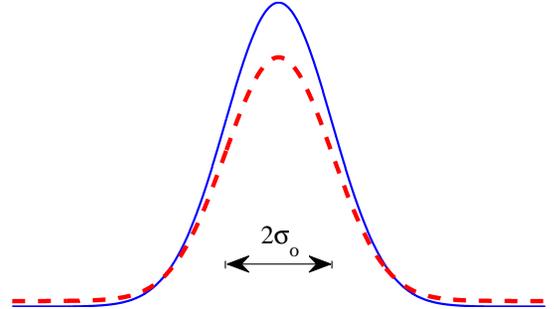

Blue: $\rho = 0$, Red: $\rho = 0.2$.

Figure 1: Plots of the measurements (as a function of $x$) for $\rho = 0$ and $\sigma_o = 1$ (top) and the density of the noise $n_t(x)$ (bottom).

$t$, the true state is the position $z_t$; the measurements correspond to a 1001-dimensional vector $\mathbf{M}(t) = [M(-500, t), M(-499, t), \ldots, M(499, t), M(500, t)]$ for locations in a grid $G = \{-500, -499, \ldots, 499, 500\}$, generated by an additive noise process

$$M(x, t) = H(x, z_t) + n_t(x).$$

Here, $H(x, z_t)$ is the square pulse function of width $2W$ around the true state $z_t$: $H(x, x_t) = 1$ if $|x - z_t| \leq W$ and 0 otherwise (see Figure 1, top). The additive noise $n_t(x)$ is randomly generated independently for each $t$ and each $x \in G$, using the mixture distribution

$$(1 - \rho) \cdot \mathcal{N}(0, \sigma_o^2) + \rho \cdot \mathcal{N}(0, (10\sigma_o)^2)$$

(see Figure 1, bottom). The parameter $\sigma_o$ represents how noisy the measurements are relative to the signal, and the parameter $\rho$ represents the fraction of outliers. In our experiments, we fix $W = 50$ and vary $\sigma_o$ and $\rho$. The total number of time steps we track for is $T = 200$.

In the generative framework, the dynamics of the object is represented by the transition model $x_{t+1} \sim \mathcal{N}(x_t, \sigma_d^2)$, and the observations are represented by the measurement process $M(x, t) \sim \mathcal{N}(H(x, x_t), \sigma_o^2)$. Thus, when $\rho = 0$, the observations are generated according to the measurement process supplied to the generative framework; for $\rho > 0$, a $\rho$ fraction of the observations are outliers.

In the online learning-based framework, the expected state dynamics function $F$ is the identity function, and the observation loss of a path $\mathbf{p} = (x_1, x_2, \ldots)$ at time $t$ is given by

$$\ell_o(x_t, M(\cdot, t)) = - \sum_{x \in [x_t - W, x_t + W] \cap G} q(M(x, t))$$

where $q(y) = \min(1 + \sigma_o, \max(y, -\sigma_o))$ clips the measurements to the range $[-\sigma_o, 1 + \sigma_o]$. That is, the observation loss for $x_t$ with respect to $M(\cdot, t)$ is the negative sum of thresholded measurement values $q(M(x, t))$ for $x$ in an interval of width $2W$ around $x_t$.

Given only the observation vectors $\mathbf{M}$, we use three different methods to estimate the true underlying state sequence $(z_1, z_2, \ldots)$. The first is the Bayesian algorithm, which recursively applies Bayes' rule to update a posterior distribution using the transition and observation model. The posterior distribution is maintained at each location in the discretization $G$. For the Bayesian algorithm, we set $\sigma_o$ to the actual value of $\sigma_o$ used to generate the observations, and we set $\sigma_d = 2$. The value of $\sigma_d$ was obtained by tuning on measurement vectors generated with the same true state sequence, but with independently generated noise values. The prior distribution over states assigns probability one to the true value of $z_1$ (which is 0 in our setup) and zero elsewhere. The second algorithm is our algorithm (NH) described in Section 4. For our algorithm, we use the parameters $\Sigma_* = 400$ and $\alpha = 0.02$. These parameters were also obtained by tuning over a range of values for $\Sigma_*$ and $\alpha$. We also compare our algorithm with the particle filter (PF), which uses the same parameters as with the Bayesian algorithm, and predicts using the expected state under the (approximate) posterior distribution. For our algorithm, we use $N = 100$ actions, and for the particle filter, we use $N = 100$ particles. For our experiments, we use an implementation of the particle filter due to (de Freitas, 2000).

Figures 2 and 3 show the true state and the states predicted by our algorithm (Blue) and the Bayesian algorithm (Red) for two different values of $\sigma_o$ for 5 independent simulations. Table 1 summarizes the performance of these algorithms for different values of the parameter $\rho$, for two different values of the noise parameter $\sigma_o$. We report the average and standard deviation of the RMSE (root-mean-squared-error) between the true state and the predicted state. The RMSE is computed over the $T = 200$ state predictions for a single simulation, and these RMSE values are averaged over 100 independent simulations.

Our experiments show that the Bayesian algorithm performs well when $\rho = 0$, that is, it is supplied with the correct noise model; however, its performance degrades rapidly as $\rho$ increases, and becomes very poor even at $\rho = 0.2$. On the other hand, the performance of our algorithm does not suffer appreciably when $\rho$ increases. The degradation of performance of the Bayesian algorithm is even more pronounced, when the noise is high with respect to the signal ($\sigma_o = 8$). The particle filter suffers a even higher degradation in performance, and has poor performance even when $\rho = 0.01$ (that is, when 99% of the observations are generated from the correct likelihood distribution supplied to the particle filter). Our results indicate that the Bayesian algorithm is very sensitive to model mismatches. On the other hand, our algorithm, when equipped with a clipped-loss function, is robust to model mismatches. In particular, our algorithm provides a RMSE value of 19.6 even under high noise ($\sigma_o = 8$), when $\rho$ is as high as 0.4.

We note that the degradation in the performance of the Bayesian algorithm is solely due to a model mismatch: when the same experiments are repeated, with the Bayesian algorithm being supplied with the correct likelihood model, it performs at least as well as, or better than our algorithm. Moreover, if the Bayesian algorithm is supplied with a likelihood model with the correct distribution (a mixture of two Gaussians), the correct fraction of outliers $\rho$, but a different outlier variance (e.g., $2\sigma_o$ instead of $10\sigma_o$), the performance of the Bayesian algorithm improves significantly over having the incorrect distribution (although it still performs worse than our algorithm).

We performed some additional experiments with our algorithm to understand the effect of varying the parameters $\Sigma_*$ and $\alpha$; the details are omitted due to lack of space. The results indicate that the performance of our algorithm depends on the value of $\Sigma_*$; if $\Sigma_*$ is set too high, there are not enough actions drawn from the regions of the state-space where the object to be tracked is, which makes it difficult to track with a fine granularity. If $\Sigma_*$ is set too low, then the actions are very concentrated at where the current actions were, and do not explore enough of the state space. On the other hand, for a low level of outliers (about 20 percent or less), our algorithm appears to have more or less the same performance over a wide range of $\alpha$ values.

## 6 RELATED WORK

The generative approach to tracking has roots in control and estimation theory, starting with the seminal work of Kalman (Kalman, 1960). The most popular generative method used in tracking is the particle fil-

Table 1: Experimental Results. Root-mean-squared-errors of the predicted positions over $T = 200$ time steps for our algorithm (NH), the Bayesian algorithm, and the particle filter (PF). The reported values are the averages and standard deviations over 100 simulations.

| Low Noise ($\sigma_o = 1$) | | | |
|---|---|---|---|
| $\rho$ | NH | Bayes | PF |
| 0.00 | 3.18 ± 0.33 | 1.17 ± 0.09 | 1.23 ± 0.11 |
| 0.01 | 3.21 ± 0.34 | 1.90 ± 0.25 | 3.98 ± 1.06 |
| 0.05 | 3.26 ± 0.34 | 3.99 ± 0.52 | 81.70 ± 1.74 |
| 0.10 | 3.31 ± 0.35 | 6.40 ± 0.84 | 81.70 ± 1.74 |
| 0.15 | 3.42 ± 0.34 | 8.38 ± 1.10 | 81.70 ± 1.74 |
| 0.20 | 3.52 ± 0.41 | 10.28 ± 1.24 | 81.70 ± 1.74 |

| High Noise ($\sigma_o = 8$) | | | |
|---|---|---|---|
| $\rho$ | NH | Bayes | PF |
| 0.00 | 10.93 ± 2.52 | 10.98 ± 2.33 | 14.35 ± 5.16 |
| 0.01 | 11.26 ± 3.43 | 12.76 ± 3.07 | 44.29 ± 16.7 |
| 0.05 | 12.03 ± 3.47 | 19.75 ± 6.70 | 81.70 ± 1.74 |
| 0.10 | 12.25 ± 2.93 | 27.33 ± 10.9 | 81.70 ± 1.74 |
| 0.15 | 13.38 ± 3.07 | 32.78 ± 13.1 | 81.70 ± 1.74 |
| 0.20 | 14.15 ± 3.88 | 43.99 ± 26.8 | 81.70 ± 1.74 |

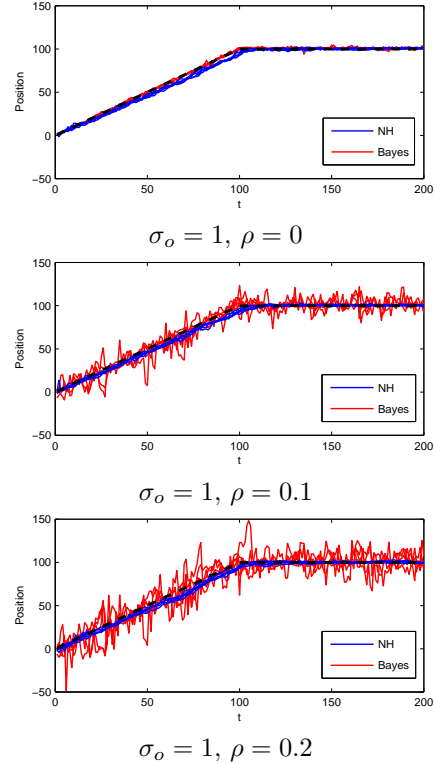

Figure 2: Predicted paths in five simulations (low noise cases). The blue lines correspond to our algorithm, the red lines correspond to the Bayesian algorithm, and the dashed black line represents the true states.

ter (Doucet et al., 2001), and its numerous variants. The literature here is vast, and there have been many exciting developments in recent years (*e.g.* (van der Merwe et al., 2000; Klaas et al., 2005)); we refer the reader to (Doucet & Johansen, 2008) for a detailed survey of the results.

The sub-optimality of the Bayesian algorithm under model mismatch has been investigated in other contexts such as classification (Domingos, 2000; Grünwald & Langford, 2007). The view of the Bayesian algorithm as an online learning algorithm for log-loss is well-known in various communities, including information theory / MDL (Merhav & Feder, 1993; Grünwald, 2007) and computational learning theory (Freund, 1996; Kakade & Ng, 2004). In our work, we look beyond the Bayesian algorithm and log-loss to consider other loss functions and algorithms that are more appropriate for our task.

There has also been some work on tracking in the online learning literature (see, for example, (Herbster & Warmuth, 1998; Koolen & de Rooij, 2008)). Our method uses a different class of actions than the tracking framework in these works. The algorithm of (Herbster & Warmuth, 1998) considers, as their class of actions, paths that switch states at most a fixed number of times. Moreover, their algorithm treats all switches between states equally, and therefore fails to take advantage of prior knowledge about the state dynamics. In contrast, we take into account such prior knowledge using a dynamics loss so that sequences approximately following the expected dynamics end up with lower loss than those that do not, and as a result, we can work with more sophisticated state transition dynamics methods.

## 7 CONCLUSIONS

In this paper, we introduce a new framework for tracking based on online learning. We propose a new algorithm for tracking in this framework that deviates significantly from the Bayesian approach. Experimental results show that our algorithm significantly outperforms the Bayesian algorithm, when the observations are generated by a distribution deviating slightly from the model supplied to the Bayesian algorithm. Our work reveals an interesting connection between decision theoretic online learning and Bayesian filtering.

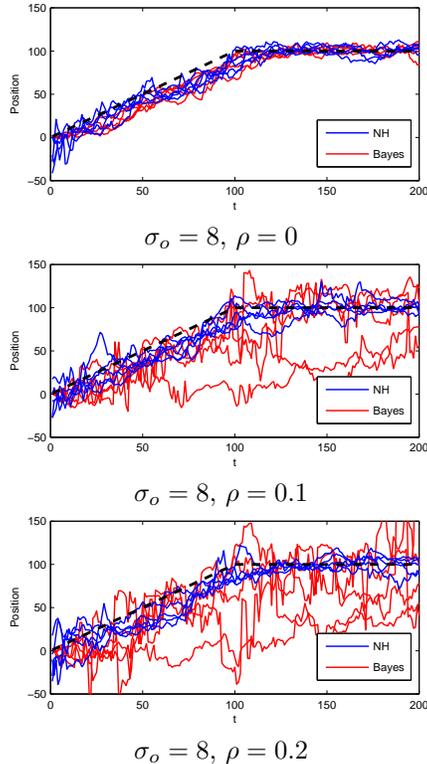

Figure 3: Predicted paths in five simulations (high noise cases).

# 8 ACKNOWLEDGEMENTS

The authors would like to thank Nando de Freitas for helpful comments. Part of this work was done when KC was part of the ITA Center at UCSD. Research support for this work was provided by NSF, under grants IIS-0713540, and IIS 0812598. KC would like to thank CALIT-2 for support. Some of the experimental results were made possible by support from the UCSD FWGrid Project, NSF Research Infrastructure Grant Number EIA-0303622.